\documentclass[journal]{IEEEtran}
\ifdefined\pdfsuppressptexinfo
\fi
\usepackage{cite}
\usepackage{amsmath,amssymb,amsfonts}
\usepackage{graphicx}
\usepackage{textcomp}
\usepackage{hyperref}
\usepackage{enumitem}
\usepackage{bm}

\begin{document}

\title{Graph-Based Recognition of Simulated Train-Driver States From Facial and Upper-Body Keypoints}

\author{Olivia~Nocentini, Marta~Lagomarsino, Gokhan~Solak, Younggeol~Cho, Qiyi~Tong, Sara~Zeynalpour, Marta~Lorenzini, Alessandro~Ledda, and~Arash~Ajoudani%
\thanks{O. Nocentini, M. Lagomarsino, G. Solak, Q. Tong, S. Zeynalpour, M. Lorenzini, and A. Ajoudani are with the Human-Robot Interfaces and Interaction Laboratory, Istituto Italiano di Tecnologia, Via S.Quirico 19d, Genoa, Italy.}%
\thanks{Y. Cho is with the Department of Bioengineering, Sir Michael Uren Hub, Imperial College London, 86 Wood Lane, London W12 0BZ, United Kingdom.}%
\thanks{A. Ledda is with the Department of Technological Innovations and Safety of Plants, Products and Anthropic Settlements, Italian National Institute for Insurance against Accidents at Work (INAIL), Via R.Ferruzzi 38, Rome, Italy.}%
\thanks{Corresponding author: Olivia Nocentini (e-mail: olivia.nocentini@iit.it).}}

\maketitle

\begin{abstract}

Driver fatigue poses a significant challenge to railway safety, with traditional systems like the dead-man switch offering limited and basic alertness checks. 
This study presents a vision-based monitoring system
that relies solely on a single front-facing RGB camera and a graph neural network to classify simulated
train-driver states into alert, not-alert, and an emergency class
comprising acted emergency-like behaviours. %

To optimize input representations for the model, an ablation study was performed, comparing three feature configurations: skeletal-only, facial-only, and a combination of both. 

Experimental results show that combining facial and skeletal features yields the highest accuracy (81\%) for the three-class model under the \textit{light} condition, outperforming models that use only facial or skeletal features. Furthermore, the combination of facial and skeletal features achieves 99\% accuracy in the \textit{alert}/\textit{not alert} classification in \textit{light} condition.
Additionally, we introduced a controlled RGB video dataset containing alert, not alert, and acted emergency-like behaviours recorded under three illumination conditions. These contributions represent a step toward passive and non-contact train-driver state recognition based on facial and upper-body dynamics.
\end{abstract}


\section{Introduction}
\label{sec:introduction}
\IEEEPARstart{F}{atigue} is a critical safety concern in railway operations, where long shifts %
and repetitive activities can significantly impair a driver's alertness \cite{filtness2017causes}. Despite regulations aimed at ensuring adequate rest for train drivers, fatigue-related incidents remain alarmingly common. According to research by the AA Charitable Trust, one in eight drivers admits to falling asleep at the wheel, while nearly two-fifths have felt so tired that they feared they might nod off \cite{reportDB}.
The risk is particularly high with modern automated trains operating on night shifts and monotonous routes \cite{avizzano2019real}. While drivers remain essential for supervision, automation reduces their active engagement for long stretches, increasing the likelihood of inattention and drowsiness \cite{fan2022types}. Fatigued drivers react more slowly to hazards and, in extreme cases, may fall asleep. 
Nighttime operation further heightens this risk because drowsiness
typically increases during late-night hours \cite{fan2022types}. %

Given these risks, effective fatigue management in railway operations is essential. A traditional safety mechanism in railway operations is the dead-man switch system, which requires the driver to press a pedal periodically to confirm alertness. If the pedal is not pressed within a specified time frame, an alarm is triggered, and the train may come to a halt. While this system ensures a basic level of vigilance, it also imposes a cognitive and physical burden on the driver, potentially leading to discomfort and operational inefficiency \cite{foot2008questions}, \cite{berdal2024towards}.
To address this problem, increasingly advanced technologies for drowsiness detection have been developed and can be grouped into three broad categories. Vehicle-based systems infer alertness by monitoring driving performance indicators such as steering behavior and braking patterns \cite{shahverdy2020driver}. Physiological-based systems use wearable sensors to capture signals like heart rate variability, muscle activity, and brainwaves \cite{peppoloni2016novel}. Finally, behaviour-based systems rely on computer vision to analyze facial expressions, eye movements, head pose, yawning frequency, and skeletal posture \cite{avizzano2019real}.

The first approach primarily relies on vehicle driving data, including steering angle, speed, and acceleration. 
However, %
the effectiveness of such methods is limited in railway operations, where drivers interact with buttons and levers intermittently rather than continuously, as with car steering, making it difficult to assess alertness based on control inputs alone. 

In contrast, physiological signal-based methods provide a more direct and accurate assessment of the driver’s physical condition. %
Despite their effectiveness, physiological measurement techniques often suffer from practical limitations. Conventional equipment often presents wearability constraints and is highly susceptible to signal noise, while medical-grade devices remain costly and less feasible for large-scale deployment. 

Given these limitations, behaviour-based monitoring has become a widely investigated alternative, as it enables passive and non-contact
observation of the driver without requiring wearable instrumentation or continuous access to vehicle-control signals.
A substantial body of research has investigated vision-based driver
fatigue and drowsiness monitoring \cite{joshi2020wild, wang2023fatigue, perkins2022challenges, maheswari2022driver, pandey2023dumodds, salem2024drowsiness}. To address the ethical and safety constraints of real-world road experiments, Wang et al. \cite{wang2023fatigue} developed a virtual driving system based on the CARLA simulation platform\cite{dosovitskiy2017carla}. This system integrates a force-feedback device to simulate real driving interactions and employs a monocular camera to capture indicators such as Eye Aspect Ratio (EAR)\cite{soukupova2016eye, cech2016real}, Mouth Aspect Ratio (MAR)\cite{sikander2018driver}, and Percentage of Eyelid Closure Over the Pupil Over Time (PERCLOS)\cite{Dinges1998PerclosAV}. Using these features, a fatigue state classification model based on Random Forest was established, achieving high classification accuracy in the reported experiments.
 Albadawi et al. \cite{albadawi2023real} developed a real-time drowsiness detection system using computer vision and machine learning to monitor operators' eye patterns via a simple webcam. The system analyzes blink patterns based on neuroscience literature to assess alertness levels. It employs machine learning models, specifically, Multilayer Perceptron, Random Forest, and Support Vector Machines (SVM), to process consecutive video frames. In their paper, the output of the network is represented by two classes (awake or asleep).
 In \cite{magan2022driver}, the authors developed a Driver Assistance System focused on detecting driver drowsiness to prevent traffic accidents. The system analyzes 60-second sequences of facial images to assess alertness. Two approaches were tested:
Recurrent and Convolutional Neural Network (R-CNN) and Deep Learning with Fuzzy Logic to classify if the driver is awake or not awake.
 In  \cite{essahraui2025real}, the authors developed a real-time, non-intrusive driver drowsiness detection (DDD) system using facial analysis and machine learning techniques. The study evaluates various algorithms, including K-Nearest Neighbors (KNN), SVM, convolutional neural networks (CNNs), and advanced computer vision models like YOLOv5, YOLOv8, and Faster R-CNN and output if the person is awake or not.
However, the aforementioned approaches mostly focus  on facial keypoints, neglecting the skeletal component, which provides valuable and stable information on train driver alertness. %

Notably, most existing studies have focused on passenger cars and freight vehicles, while research on fatigue detection in railway transportation remains relatively limited. However, the conditions and challenges of alertness detection for train drivers differ from those in road transportation. Unlike car drivers, who maintain relatively stable postures with their hands on the steering wheel, train operators are seated on spring-mounted chairs that introduce head oscillations due to track-induced vibrations. They also perform wider head movements, such as scanning long stretches of track, checking side panels and control screens, and monitoring signals outside the cabin. Compared with road-vehicle drivers, train operators exhibit different
movement amplitudes and temporal patterns because they periodically
inspect the track, side panels, control displays, and external signals
rather than continuously steering a vehicle.

Among the relatively limited studies addressing alertness monitoring in railway transportation, Avizzano et al.~\cite{avizzano2019real} developed an embedded computer vision system for train driver drowsiness detection. The system relies on a single camera installed in the cockpit to monitor gaze direction and eyelid blinking as visual indicators of reduced alertness.
Moreover, another work, \cite{chen2023research}, introduced a multi-feature fatigue detection method designed for railway dispatchers, combining facial keypoints with body posture analysis to improve detection accuracy, even under challenging conditions such as facial occlusion or angle variations. This approach considers features such as eye closure duration, blinking patterns, yawning frequency, and posture-related behaviors to assess fatigue levels. The model employs an adaptive Bi-LSTM-SVM algorithm to integrate facial and behavioral data, outputting a three-level classification of fatigue.\\
Beyond fatigue-oriented monitoring, Hayashi et al. \cite{hayashi2020development} proposed a multimodal
abnormal-sign detection system for professional road drivers. Their
framework combines body and facial images with an ear-mounted pulse
sensor, steering-angle measurements, and voice interaction. Separate
modules detect a predefined set of cue signs, including touching the
head or chest, one-arm hanging, prolonged eye closure, abnormal heart
rate, and absence of steering input. The type, duration, and recurrence
of the detected cues are subsequently aggregated through a rule-based
confidence level. Depending on this confidence level, the system can
continue monitoring the driver, initiate a voice interaction to obtain
additional information about the driver's internal state, or trigger an
intervention. 
Its system-level quantitative
validation mainly addressed sleepiness, whereas pseudo-headache was
presented as an illustrative simulated scenario rather than as a
clinically validated medical condition.

Their work demonstrated that driver monitoring has
already been extended beyond conventional fatigue detection to include
behavioural and physiological cues potentially related to medical
impairment. However, %
their system relies on multiple inputs, including visual
data, a contact-based physiological sensor, vehicle-control information,
and active voice confirmation. Although this multimodal configuration
can provide richer information about the driver's internal condition,
it requires physical sensor placement, access to vehicle signals, and
integration of multiple acquisition and interaction modules.

In contrast, the present work investigates a complementary design
trade-off by prioritising passive, non-contact, and non-wearable
vision-based monitoring that relies on both facial and skeletal features. The proposed framework requires only a
single front-facing RGB video stream and does not depend on wearable
physiological sensors, train-control telemetry, or an explicit verbal
response from the operator. Therefore, no device needs to be attached
to the train driver during operation. This reduced sensing burden may
facilitate the integration of the monitoring system into existing train
cabins while limiting interference with the operator's normal activities.
Moreover, rather than detecting independently defined cue signs and combining
them through handcrafted rules, the proposed method directly learns
the joint spatiotemporal evolution of facial and upper-body landmarks
within a unified graph representation.

Accordingly, the remaining research gap is the limited
availability of railway-specific approaches that perform direct state
recognition using only non-contact visual observations. Moreover, existing
railway-monitoring methods predominantly address binary
awake--asleep classification or different levels of fatigue. Limited attention has
therefore been devoted to passive train-driver monitoring systems that
jointly model facial and upper-body dynamics and distinguish reduced
alertness from acted emergency-like behaviours using a single RGB
video stream.

A further challenge is introduced by the operating conditions of railway
vehicles. Train drivers may be exposed to abrupt illumination changes
when entering or leaving tunnels, moving between outdoor and
underground sections, or operating during nighttime conditions. Such
variations can affect the reliability of facial and skeletal landmark
estimation and, consequently, the performance of downstream
vision-based state-recognition models. This issue is particularly
relevant for a vision-only framework and therefore requires explicit
evaluation under both stable and changing illumination conditions.

Motivated by these considerations, this paper presents a graph-based
vision framework for recognising train-driver states from the joint
temporal evolution of facial and upper-body landmarks. The problem is
formulated as a three-class classification task comprising
\textit{alert}, \textit{not alert}, and \textit{emergency}
states. The latter class includes controlled behaviours such as
pretending to faint, holding the head, or clutching the chest.

The main contributions of this work are summarised as follows:

\begin{enumerate}
    \item We propose a passive, non-contact, non-wearable, and vision-only
train-driver monitoring framework based on a single front-facing RGB
video stream and relying on a graph neural network. The proposed system requires neither physiological
instrumentation, train-control telemetry, nor active responses from the
operator.

    \item We formulate train-driver monitoring as a direct three-class
state-recognition problem comprising \textit{alert}, \textit{not alert},
and an acted emergency-like class.

    \item We introduce a preliminary controlled RGB video dataset involving
    25 participants and three driver-state classes recorded under
    three railway-relevant illumination settings. The changing-illumination
    condition was designed to reproduce the visual variations that may
    occur when a train enters or exits a tunnel.

    \item We evaluate the complementary contribution of facial and
    upper-body information through an ablation study comparing
    facial-only, skeletal-only, and combined face-and-upper-body graph
    representations in the three-class scenario. In addition, an occlusion-sensitivity analysis is
    performed to investigate which facial and skeletal landmarks have
    the greatest influence on the model's predictions.
\end{enumerate}

\begin{figure*}[!t]
\centering
\includegraphics[width=1.5\columnwidth]{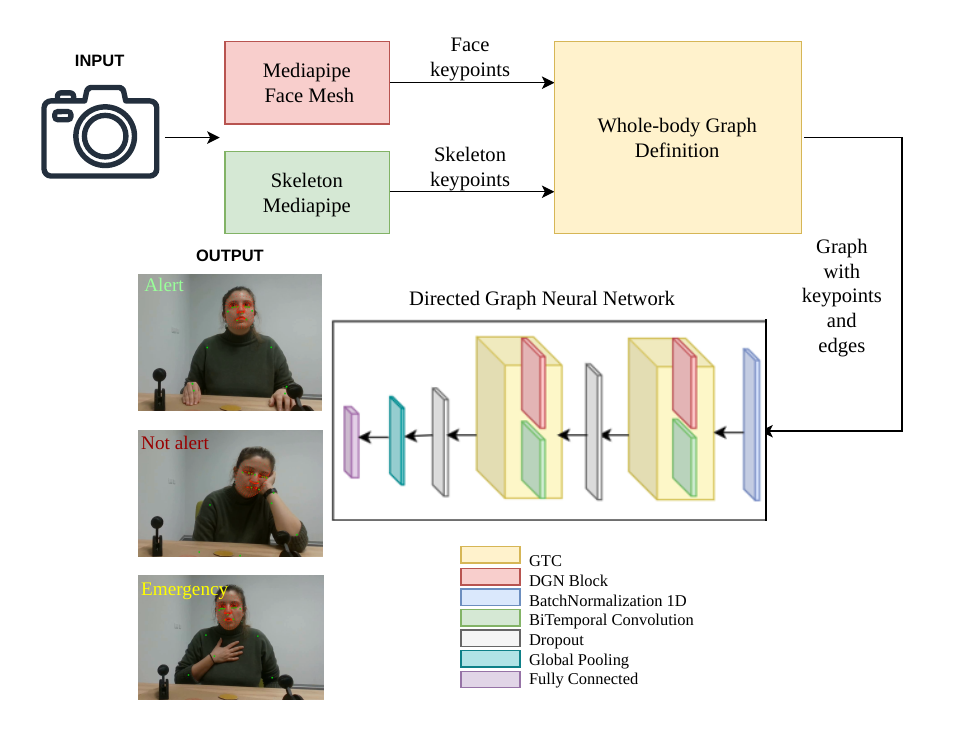}
\caption{Flowchart of our monitoring system framework. The inputs are the images acquired by an RGB camera, from which facial and skeletal features are extracted. These keypoints are used to construct a whole-body graph, which is then processed by a custom DGNN to classify the user's state into three categories: \textit{alert}, \textit{not alert}, or potentially experiencing an \textit{emergency} condition.}
\vspace{-0.5cm}
\label{flowchart}
\end{figure*}

\section{Methodology}
\label{sec:methodology}
Figure \ref{flowchart} illustrates the overall flowchart of our graph-based monitoring framework for online train driver state assessment, analyzing whole-body \footnote{Throughout this paper, the term
\textit{whole-body} is used to present the combination of facial and skeletal features.} motion only using RGB images. %
Facial and skeletal keypoints are extracted from each video frame to construct a graph. This graph is then processed by a custom Directed-Graph Neural Network (DGNN), which classifies the driver's state into one of three categories: \textit{alert}, \textit{not alert}, or potentially experiencing an \textit{emergency} condition. 

We collected a preliminary custom dataset of RGB videos simulating train driver operations under three common railway lighting conditions: \textit{light}, \textit{dark}, and \textit{sudden light/dark} (see Section \ref{sec:sub:dataset} for more details). This dataset enabled us to train the network on more realistic scenarios and conduct an ablation study alongside a model explainability analysis. These analyses aim to understand the relevance of different facial and skeletal keypoints in the network's decision process and assess its sensitivity to keypoint occlusion. 

This paper proposes a rigorous pipeline to address the following research questions (RQs): 
\begin{enumerate}[label=\textbf{RQ\theenumi}.]
    \item \label{RQ1} \textit{Can a vision and graph-based neural network effectively detect simulated train driver alertness and emergency states?
    }  
    \item \label{RQ2}  
    \textit{%
    Does considering whole-body movements improve driver state detection performance compared to using only facial or skeletal keypoints in the three-class scenario?}  
\end{enumerate}
\subsection{Estimation of facial and skeletal keypoints in video captures and whole-body graph definition}
From each video frame, we estimated the position of the upper body, employing the first 17 keypoints of Mediapipe. %
We took only the upper part of the skeleton of Mediapipe since the train driver is usually seated on a chair in front of a dashboard.
Since Mediapipe provides only few facial keypoints (corresponding to the positions of the left and right eyes, the ears, and the mouth), we integrated an additional 478 facial keypoints from Mediapipe's Face Mesh \cite{jakhete2024comprehensive}. This allowed us to better capture facial contours, eyebrows, eye openness, and mouth movements, which the skeleton of Mediapipe alone cannot detect.\\
We selected the skeleton of Mediapipe and Mediapipe Face Mesh for features extraction because both are stable feature detectors for what concerns RGB images \cite{kim2023human}, \cite{dong2024enhanced}. Moreover, since some keypoints were not consistently detected during the extraction process from the RGB videos, a masking strategy was introduced to handle missing keypoints.

\begin{figure}[!t]
\centering
\includegraphics[width=\columnwidth]{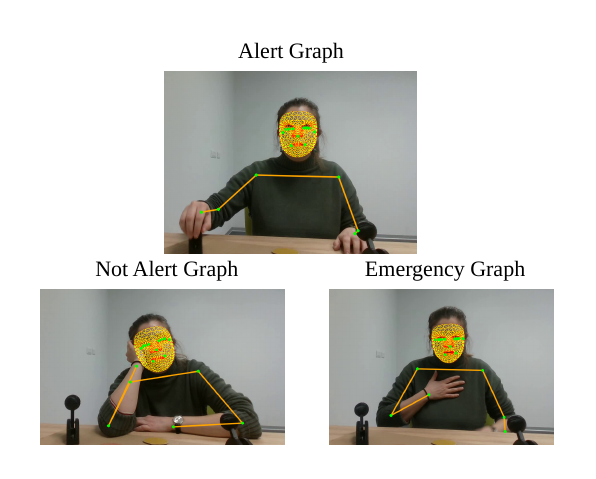}
\caption{Samples of the defined graphs, which correspond to the input of our network, in the three tested users' states: \textit{alert}, \textit{not alert}, and simulated \textit{emergency} condition. The joints of Mediapipe Face Mesh are shown in red and  the Mediapipe skeleton joints are drawn in green. The bones connecting the joints coming from Mediapipe Face Mesh are drawn in yellow, while the other edges related to the skeleton are drawn in orange.}
\vspace{-0.5cm}
\label{fig:graph}
\end{figure}

Once the keypoints are extracted, we construct a graph to represent the whole-body posture and movements of the user. More specifically, facial and upper-body keypoints, represented by their 2D coordinates $(x,y)$, serve as graph nodes, while bones define the edges that connect them. The keypoints are normalized with respect to the nose position across all three scenarios (face-only, skeleton-only, and face-and-skeleton keypoints). 
In Figure \ref{fig:graph}, samples of the graph defined by connecting the normalized keypoints through edges are reported for the three tested states: \textit{alert}, \textit{not alert}, and the simulated \textit{emergency}. These graphs represent the input of our network.

\subsection{Neural Network for Alert, Not Alert, and Emergency State Classification}
\subsubsection{Network architecture}
The proposed architecture is a customised Directed Graph Neural Network (DGNN) that took inspiration from \cite{shi2019skeleton}, designed to process structured spatiotemporal data by jointly modeling graph-based dependencies and temporal dynamics.

The model consists of two Graph-Temporal Convolutional (GTC) blocks, each integrating a Directed Graph Network (DGN) block for relational learning and a Bi-Temporal Convolution (BiTC) layer for temporal feature extraction. Each DGN block updates node and edge features based on directed graph connectivity matrices, %
using learned transformations to aggregate information from incoming and outgoing edges. This operation allows the model to capture structured interactions within the graph. The Bi-Temporal Convolution (BiTC) layer then applies temporal convolutions along the time dimension to both node and edge features, enhancing the model’s ability to learn temporal dependencies.

The input data undergoes an initial batch normalization step, separately for node and edge features, ensuring stable training. The first GTC block expands the feature dimension, while the second block applies strided temporal convolutions (stride=2), reducing the temporal resolution and enabling hierarchical feature extraction. A dropout layer (p=0.3) is applied after each GTC block to mitigate overfitting.

At the final stage, global pooling is applied to the learned node and edge features, followed by a fully connected layer that generates the output predictions. The model employs LeakyReLU activation throughout and uses a Label Smoothing Loss to enhance generalization and prevent overconfidence in predictions.

Our customised model differs from the original implementation of \cite{shi2019skeleton} in several key aspects, aimed at improving performance. First, we reduced the number of convolutional layers, creating a more compact architecture tailored to the relatively small size of our dataset. Second, we introduced dropout layers at specific stages to effectively mitigate overfitting. Third, we replaced certain ReLU activation functions with LeakyReLU, allowing the model to better handle negative values and ensuring greater training stability. Additionally, we incorporated a label smoothing loss function (LabelSmoothingLoss) to enhance robustness when training on noisy datasets, such as ours, which contains inaccuracies in the detected keypoints due to varying lighting conditions during recording.
\subsubsection{Network hyperparameters}
The customised DGNN was trained using the following hyperparameters: a learning rate of 0.001, 50 epochs, a batch size of 512, Adam as the optimizer, Label Smoothing of 0.1, and a lookback (i.e., how many frames the neural network \textit{looks back} into the past) of 30. The hyperparameters were empirically tuned based on validation-set performance and inspection of the validation curves.
\subsection{Relevance of facial and skeletal keypoints in model decision-making}
\subsubsection{Ablation study}
We conducted an ablation study to assess the impact of different categories of keypoints on the performance of the proposed DGNN for train driver's state monitoring. Specifically, we trained three network variants using different input graphs, including: %
(1) whole-body keypoints (i.e., combined facial and skeletal keypoints, to examine their synergistic effect); (2) %
facial keypoints only, to determine whether the facial expressions and head movements alone suffice for state estimation, as commonly done in the literature; and (3) skeletal keypoints only, to investigate the importance of upper-body posture and movements in isolation. This approach allowed us to isolate the contributions of facial and skeletal keypoints and determine how their integration enhances the model’s performance. 
\subsubsection{Occlusion Sensitivity Analysis}
This analysis aims to identify the most significant keypoints for the functioning of the proposed DGNN and accurate prediction generation. To achieve this, we employed the Occlusion Sensitivity Analysis, a method for understanding the importance of individual features for a deep network's classification. The technique involves occluding (i.e., hiding or removing) one or more keypoints from the input data and observing how this alteration affects the model's performance or predictions. If the occlusion of a specific keypoint causes a significant degradation in the model's predictions, it can be concluded that the keypoint is particularly important for the network's decision-making process.

\section{Experimental Campaign}
\label{sec:experiments}
This section details the experimental setup, the dataset, and the analysis performed.
\subsection{Setup}
\begin{figure}[!t]
\centering
\includegraphics[width=0.8\columnwidth]{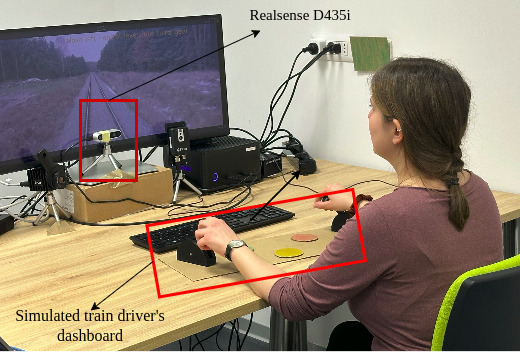}
\caption{The experimental setup consisted of users seated at a desk, following on-screen instructions to manipulate control levers, press dashboard buttons, and check side panels. An RGB camera continuously monitored the subject from the front. 
}
\vspace{-0.5cm}
\label{setup}
\end{figure}
The experimental setup (Fig. \ref{setup}), involved participants seated on a chair in front of a table equipped with a computer monitor and a Realsense D435i camera. This camera was positioned directly in front of the participant to capture RGB videos for dataset collection. The sensor operated at a frame rate of 30 fps, producing images with a resolution of 640$\times$480 pixels.
\subsection{Dataset} %
\label{sec:sub:dataset}
The artificial dataset, used for preliminary validation of our framework, consists of videos carried out at Human-Robot Interfaces and Interaction (HRII) Lab, Istituto Italiano di Tecnologia (IIT), in accordance with the Helsinki Declaration. The experimental protocol was approved by the ethics committee Azienda Sanitaria Locale Genovese N.3 (Protocol IIT\_HRII\_ERGOLEAN 156/2020) and participants were recruited from the student body and research staff at IIT. The dataset was divided into training, validation, and test sets, comprising data from 15, 5, and 5 subjects, respectively. In total, the dataset included 25 participants, consisting of 14 males and 11 females, with ages ranging from 23 to 36 years. Of these participants, 12 wore glasses, while 13 did not.

For each user, we acquired 9 videos of two minutes and half each: the first 3 of them in which the subject is \textit{alert}, the second 3 in which the subject is \textit{not alert}, and the last 3 when the user is pretending to experience an \textit{emergency} condition. Each condition (\textit{alert}, \textit{not alert} and \textit{emergency}) is recorded in 3 different lighting conditions categorized as \textit{light},  \textit{dark}, and \textit{sudden light/dark}, according to the intensity of the light created in the room.
The \textit{light} condition was defined as the room with 4 lights on, the \textit{dark} condition was defined as the room without light, and the \textit{sudden light/dark} condition was when the lights were turned on/off every 15 seconds to simulate entering/exiting a tunnel.\\
During the \textit{alert} phase, the subject was asked to watch a video created by the authors, depicting a train track view. This was designed to help the participant immerse themselves in the role of a train driver. Additionally, instructions appeared on the screen, guiding the participant to perform specific actions, such as moving levers, pressing buttons on the dashboard, or checking side panels. These prompts encouraged head and body movements, enhancing the system’s robustness to variations in head orientation and positioning relative to the camera.\\
During the \textit{not alert} phase, subjects were required to keep their eyes closed and pretend to be asleep.
Finally, during the \textit{emergency} phase, participants simulated experiencing a medical, as if suffering from a sudden illness. The emergency class does not represent clinically validated medical events. Rather, it consists of acted emergency-like behaviours (i.e pretending to faint or clutching their chest or putting both of their hands on the head) designed to provide a preliminary proxy class for abnormal driver conditions in a controlled experimental setting.
Importantly, across both the \textit{not alert} and \textit{emergency} phases, some subjects occasionally covered parts of their faces with their hands, creating self-occlusions. This added variability to the cues and made the scenarios more realistic by reflecting real conditions where facial features may be obscured.

\subsection{Analysis}
\subsubsection{Accuracy}
Accuracy during neural network training measures the proportion of correct predictions to the total number of samples, calculated as:
\begin{equation}
\text{Accuracy} = \frac{\text{Number of Correct Predictions}}{\text{Total Number of Samples}} \times 100 \%.
\end{equation}
It evaluates how well the model predicts the true class (ground truth) across the dataset, providing a straightforward metric for performance assessment. Accuracy ranges from 0 to 100\% and is particularly useful for classification tasks with balanced class distributions.
\subsubsection{Confusion Matrices}
We computed the confusion matrices (i.e., a performance evaluation tool comparing predictions to actual labels) of the proposed DGNN using whole-body keypoints for state monitoring in three different lighting conditions for three and two class classification.
Moreover, we computed the confusion matrices also for the only skeleton and only facial scenarios.

\section{Results}
\label{sub:results}

\begin{figure*}[!t]
\centering
\includegraphics[width=1.6\columnwidth]{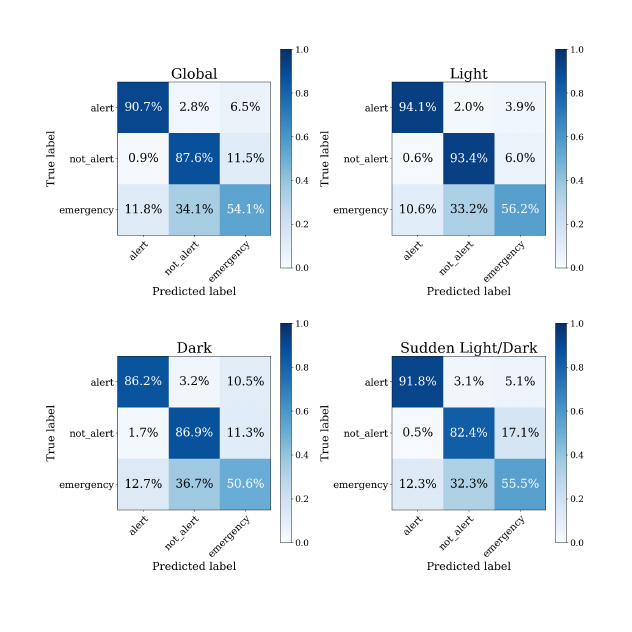}
\caption{\centering Confusion Matrices for the whole-body graph in the three-class conditions.
}
\vspace{-0.5cm}
\label{cm}
\end{figure*}

To address the  \ref{RQ1} and \ref{RQ2}, we computed the metrics on the 5 test subjects ( accuracy, precision, recall and F1-Score) of the proposed DGNN model for three-class classification. %
The results using whole-body, facial-only, and skeletal-only keypoints are illustrated in Tables  \ref{table_fs}, \ref{table_f}, \ref{table_s}, while the performance of the alert/not alert classes is presented in Table \ref{table_fs_2c}.%
The best performance is observed when only two classes are considered (\textit{alert}/\textit{not alert}), with the test accuracy of 99\%. 
The inclusion of the \textit{emergency} class reduces accuracy, but the three-class whole-body model still performs well, outperforming models that rely solely on facial or skeletal keypoints. %
After the whole-body features, the best results are achieved using facial-only features, while skeletal-only features obtain the lowest performance.

Moreover, the performance is highest in the full \textit{light} condition, followed by the variable \textit{sudden light/dark} condition in all the scenarios.

To better understand class misclassifications under various lighting conditions, we analyzed the confusion matrices for our three-class model using facial and skeletal keypoints (see Figure \ref{cm}). The analysis revealed that the \textit{emergency} state is sometimes misclassified as the \textit{not alert} condition.\\
A video providing supplementary material for the paper, with a focus on the inference online, is available at this link:
\url{https://youtu.be/Cm-8flfn91E}.\\
Finally, to further address \ref{RQ2} on the individual and combined contribution of facial and skeleton keypoints to driver state estimation, we conducted an Occlusion Sensitivity Analysis on the proposed three-class whole-body model. The results, illustrated in Figures \ref{occlusion_total}, show the relative importance of each keypoint for correct classification. Each bar in the plot represents a keypoint, with its height indicating its significance in distinguishing between the three driver states.

\begin{table*}
\centering
\caption{\centering Whole-body features metrics in general, light, dark, and sudden light/dark lighting conditions for the three class scenario}
\begin{tabular}{|l|l|l|l|l|}
\hline
\textbf{Light Condition} & \textbf{Accuracy (\%)} & \textbf{Precision (\%)} & \textbf{Recall (\%)} & \textbf{F1-Score (\%)} \\ \hline
Global                  &       78                 &     78                    & 77                     &        78                \\ \hline
Light                    &             81          &   82                     &            81       &              80          \\ \hline
Dark                     &     75                   &       75                 &             75         &         74               \\ \hline
Sudden Light/Dark        &   77                     &          76               &            77          &               76         \\ \hline
\end{tabular}
\label{table_fs}
\end{table*}

\begin{table*}
\centering
\caption{\centering Facial-only features metrics in general, light, dark, and sudden light/dark lighting conditions for the three class scenario}
\begin{tabular}{|l|l|l|l|l|}
\hline
\textbf{Light Condition} & \textbf{Accuracy (\%)} & \textbf{Precision (\%)} & \textbf{Recall (\%)} & \textbf{F1-Score (\%)} \\ \hline
Global                  &       75               &            75           &                  75   &        75            \\ \hline
Light                    &      78                  &             78            &             78         &          77              \\ \hline
Dark                     & 76                       &       75                  &              76        &        75                \\ \hline
Sudden Light/Dark        &        72                &      71                   &                 72     &        72                \\ \hline
\end{tabular}
\label{table_f}
\end{table*}

\begin{table*}
\centering
\caption{\centering Skeletal-only features metrics in general, light, dark, and sudden light/dark lighting conditions for the three class scenario}
\begin{tabular}{|l|l|l|l|l|}
\hline
\textbf{Light Condition} & \textbf{Accuracy (\%)} & \textbf{Precision (\%)} & \textbf{Recall (\%)} & \textbf{F1-Score (\%)} \\ \hline
Global                &         61               &        66                 &               61       &             60           \\ \hline
Light                    &          64              &          68               &            64          &              64          \\ \hline
Dark                     &      55                  &     61                    &           55           &      54                  \\ \hline
Sudden Light/Dark        &           63             &           68              &              63       &           63             \\ \hline
\end{tabular}
\label{table_s}
\end{table*}

\begin{table*}
\centering
\caption{\centering Whole-body  features metrics in general, light, dark, and sudden light/dark lighting conditions for the two-class scenario}
\begin{tabular}{|l|l|l|l|l|}
\hline
\textbf{Light Condition} & \textbf{Accuracy (\%)} & \textbf{Precision (\%)} & \textbf{Recall (\%)} & \textbf{F1-Score (\%)} \\ \hline
Global                &           97             &        97                 &              97        &           97             \\ \hline
Light                    &    99                    &       99                  &               99       &    99                    \\ \hline
Dark                     &   94                     &          94               &                94      &          94              \\ \hline
Sudden Light/Dark        &      96                  &        96                 &             96         &    96                    \\ \hline
\end{tabular}
\label{table_fs_2c}
\end{table*}

\section{Discussion}
The results of this study highlight that a graph-based neural network using whole-body keypoints analysis was highly effective for binary driver alertness classification and achieved an overall three-class accuracy of 78\% for distinguishing among three classes: \textit{alert}, \textit{not alert}, and simulated \textit{emergency} condition. %
Instead, using only skeletal or facial features was insufficient for appropriate three-class classification %
using DGNN, particularly under the varying lighting conditions. 
\par
The row-normalized confusion matrices provide further insight into the classification behaviour of the proposed model across the different lighting conditions. Overall, the model shows a strong ability to discriminate between the \textit{alert} and \textit{not alert} classes. In the global setting, 90.7\% of the \textit{alert} samples and 87.6\% of the \textit{not alert} samples are correctly classified, with only limited confusion between these two classes. This suggests that the model is able to effectively capture discriminative behavioural patterns associated with normal alertness and reduced alertness.

In contrast, the \textit{emergency} class represents the most challenging condition to identify. In the global scenario, only 54.1\% of the emergency samples are correctly classified, while 34.1\% are misclassified as \textit{not alert}.  This misclassification probably occurred because, during data collection, several participants adopted similar whole-body postures in both conditions. For instance, some subjects slumped forward or tilted their heads
backwards both when experiencing an emergency event and when falling asleep, making it challenging for the network to distinguish between the two conditions.
A key limitation of this study is that both the \textit{not alert} and \textit{emergency} classes were simulated in a controlled laboratory setup. Therefore, the proposed framework should be interpreted as a preliminary state-classification system for acted scenarios, not as a clinically validated detector of real fatigue or medical emergencies. Future work will require in-cab data, expert-informed scenario design, and real-world validation to assess operational generalization.
Furthermore, the misclassification between \textit{emergency} and \textit{not alert} explains why the test accuracy of the two-class model was higher compared to the three-class model (see Tables \ref{table_fs_2c} and Table \ref{table_fs}).

When considering the different illumination settings, the best performance is obtained in the \textit{light} condition, where both \textit{alert} and \textit{not\ alert} classes achieve high recognition rates, equal to 94.1\% and 93.4\%, respectively. The \textit{emergency} class also reaches its highest value in this condition, with 56.2\% of correctly classified samples. Conversely, the \textit{dark} condition leads to the lowest performance, particularly for the \textit{emergency} class, which decreases to 50.6\%. This is because RGB cameras work well under well-lit conditions rather than in darkness. This can be explained not only by the intrinsic limitations of RGB cameras in low-light environments, but also by the fact that the keypoint extraction stage relies on Mediapipe models trained mainly on RGB datasets collected under favorable illumination conditions. Consequently, keypoint detection becomes less reliable in dark scenes, which negatively impacts the downstream classification performance. Future works will investigate the use of thermal cameras to enhance the performance of the network in dark conditions.

Finally, in the \textit{sudden light/dark} condition, the model maintains good performance for the \textit{alert} class, correctly classified in 91.8\% of cases. However, a higher confusion is observed between the \textit{not  alert} and \textit{emergency} classes, with 17.1\% of \textit{not alert} samples misclassified as emergency and 32.3\% of emergency samples misclassified as \textit{not alert}. This suggests that abrupt illumination changes may introduce additional variability in the extracted features, making the distinction between reduced alertness and emergency states more difficult.
\par
\begin{figure}[!t]
\centering
\includegraphics[width=\columnwidth]{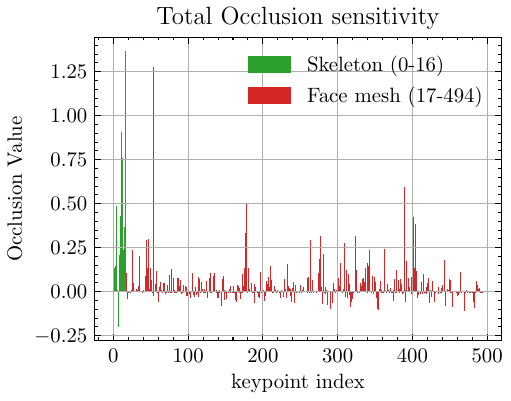}
\caption{\centering \centering Results of the Occlusion Sensitivity Analysis revealing the
significance of individual keypoints among the three classes for the whole-body configuration.
}
\vspace{-0.5cm}
\label{occlusion_total}
\end{figure}

The Occlusion Sensitivity analysis showed that in  the overall ranking (Figure \ref{occlusion_total}), skeletal keypoints were among the most influential features, particularly keypoint 16, corresponding to the right wrist, followed by keypoints 11 and 12, corresponding to the left and right shoulders, as well as keypoints 9 and 10, corresponding to the mouth region, and keypoint 15, corresponding to the left wrist. This suggests that the model strongly exploits posture-related information, especially shoulder configuration and hand/wrist position, together with facial-orientation or mouth-related cues, to discriminate the operator’s state. 
In conclusion, these findings indicate that skeletal landmarks related to shoulders and wrists play a central role in the model’s decision process, while facial landmarks provide complementary and more localized information.\\
Regarding the impact of topology, the Occlusion Sensitivity Analysis provides information about the importance of individual nodes but does not directly isolate the contribution of alternative edge configurations. A systematic comparison among different graph
topologies, root definitions, and connectivity patterns was not performed in the present study and would
require a dedicated ablation analysis. We therefore identify topology comparison as a specific direction for
future work.

\par
A direct quantitative comparison with existing approaches was not considered methodologically appropriate because the available studies generally rely on different datasets, acquisition protocols, class definitions, and evaluation procedures. In particular, most previous driver-monitoring approaches focus on binary \textit{alert}/\textit{not alert} classification and are evaluated under a day/night or relatively stable illumination condition, whereas the present study considers three behavioural states and multiple challenging lighting scenarios. Consequently, the reported results are not directly comparable with those in the literature.

Future works could explore Graph Attention Networks (GATs) or Vision Transformers (ViTs) to enhance feature selection by learning an adaptive attention map, focusing on the most discriminative facial and skeletal features for each state. Incorporating these techniques could lead to more robust and explainable human behavior recognition using only vision sensors, improving state classification accuracy while reducing reliance on redundant keypoints. Finally, we plan to expand the dataset (increasing robustly the number of subjects involved in the study) and include occlusions (hats, masks) and real-world artifacts like camera vibrations from train movement.

\section{Conclusions}
Driver fatigue remains a critical challenge in railway safety, yet traditional monitoring systems, such as the dead-man switch, offer only rudimentary alertness checks. In this study, we proposed a vision-based monitoring system leveraging a customised Directed-Graph Neural Network (DGNN) and whole-body movement analysis to classify train driver states into \textit{alert}, \textit{not alert}, and simulated \textit{emergency} conditions. %
To determine the most effective input representation for the network, we conducted an ablation study comparing three different keypoint configurations: skeletal-only, facial-only, and whole-body graph. Results indicate that the whole-body graph yields the highest performance, achieving a test accuracy of 81\% for the three-class classification in the \textit{light} condition. %
For the two-class classification, the accuracy is 99\% in \textit{light} condition.

Furthermore, we introduced a controlled vision-based dataset containing
alert, not-alert, and acted emergency-like behaviours in simulated
train-driver scenarios.
This study deliberately focused on standard RGB sensing to investigate
a low-complexity and widely deployable vision-only configuration.
Future work will examine complementary thermal or near-infrared sensing
to improve robustness under low-light conditions and will expand the acquired dataset. Moreover, we will leverage GATs or ViTs to enhance feature selection through adaptive attention maps and focus on the most relevant facial and body keypoints. This approach could improve classification accuracy, provide more interpretable human behavior recognition, and reduce reliance on redundant keypoints.

\section*{Acknowledgment}
This work was supported by the National Institute for Insurance against Accidents at Work INAIL within the VIVA project.

\bibliographystyle{IEEEtran}
\bibliography{ref}

\end{document}